\documentclass[runningheads,orivec]{llncs}
\usepackage{esvect}
\usepackage[T1]{fontenc}
\usepackage{graphicx}
\usepackage{graphicx}
\usepackage{booktabs}
\usepackage{amsmath}
\usepackage{amssymb}
\usepackage{multirow}
\usepackage{array}
\usepackage{makecell}
\usepackage{amsfonts}
\usepackage{algorithm}
\usepackage{algpseudocode}
\usepackage{setspace}
\usepackage{enumitem}
\usepackage{bbm}
\usepackage{bm}
\usepackage[dvipsnames]{xcolor}
\usepackage{pifont}
\usepackage{courier}
\usepackage{wrapfig}
\usepackage{tabularx}
\usepackage{hyperref}
\usepackage{color}

\hypersetup{
    colorlinks=true,
    linkcolor=blue,
    citecolor=blue,
    filecolor=blue,
    urlcolor=blue
}
\begin{document}
\title{SCALE: Semantic- and Confidence-Aware Conditional Variational Autoencoder for Zero-shot Skeleton-based Action Recognition}
\titlerunning{SCALE}
%
\author{Soroush Oraki\inst{1} \and
Feng Ding\inst{1} \and
Jie Liang\inst{2}\vspace{3mm}\thanks{Corresponding author.}}
\authorrunning{S. Oraki et al.}
%
\institute{School of Engineering Science, Simon Fraser University, Burnaby, BC, Canada\\
\email{\{soa32,fda17\}@sfu.ca} \and
School of Information Science and Technology, Eastern Institute of Technology, Ningbo, China\\
\email{jliang@eitech.edu.cn}
}
\maketitle              
\begin{abstract}
Zero-shot skeleton-based action recognition (ZSAR) aims to recognize action classes without any training skeletons from those classes, relying instead on auxiliary semantics from text. Existing approaches frequently depend on explicit skeleton–text alignment, which can be brittle when action names underspecify fine-grained dynamics and when unseen classes are semantically confusable. We propose SCALE, a lightweight and deterministic Semantic- and Confidence-Aware Listwise Energy-based framework that formulates ZSAR as class-conditional energy ranking. SCALE builds a text-conditioned Conditional Variational Autoencoder where frozen text representations parameterize both the latent prior and the decoder, enabling likelihood-based evaluation for unseen classes without generating samples at test time. To separate competing hypotheses, we introduce a semantic- and confidence-aware listwise energy loss that emphasizes semantically similar hard negatives and incorporates posterior uncertainty to adapt decision margins and reweight ambiguous training instances. Additionally, we utilize a latent prototype contrast objective to align posterior means with text-derived latent prototypes, improving semantic organization and class separability without direct feature matching. Experiments on NTU-60 and NTU-120 datasets show that SCALE consistently improves over prior VAE- and alignment-based baselines while remaining competitive with diffusion-based methods.

\end{abstract}
%
%

%
%
\section{Introduction}

Human action recognition (HAR)~\cite{kong2022human,kumar2024survey,wang2023comprehensive} aims to identify human activities from observed motion and has become a core capability for applications such as surveillance~\cite{khan2024human,kamthe2018suspicious,singh2019multi}, human computer interaction~\cite{gammulle2023continuous,haria2017hand}, and healthcare systems~\cite{SkeletonFall,ramirez2021fall}. While RGB videos have historically driven progress in HAR due to data availability, they also inherit substantial nuisance factors (background clutter, illumination, viewpoint changes) and raise privacy concerns. Motivated by advances in depth sensors and pose estimation~\cite{dibenedetto2025comparing}, skeleton-based action recognition has emerged as an efficient alternative that represents human motion as compact 3D joint trajectories, offering robustness to appearance variations, lighting, and different camera angles, as well as improved privacy preservation.

Despite strong supervised performance~\cite{stgcn,do2024skateformer,cheng2020skeleton,duan2022revisiting,ding2025lstc,chen2021channelwise,oraki2024lortsar,Chi_2022_CVPR,zhou2024blockgcn}, fully supervised skeleton-based models typically require exhaustive annotated datasets and often struggle to generalize beyond the closed set of training actions. This limitation is particularly restrictive in open-world settings, where new actions continuously appear and collecting labeled skeleton sequences for every class is impractical. Zero-shot skeleton-based action recognition (ZSAR)~\cite{zhu2024part,gupta2021syntactically,hubert2017learning,cada,li2024sa,zhou2023zero,chen2024fine,tdsm} addresses this challenge by recognizing unseen action classes using auxiliary semantic information, commonly text descriptions derived from action labels. The central premise is that unseen actions share transferable motion primitives with seen ones, and that language can provide a semantic bridge for transferring knowledge.

Recent ZSAR methods primarily focus on cross-modal alignment between skeleton features and text embeddings. Early efforts learn a shared embedding space or reduce distributional mismatch across modalities~\cite{hubert2017learning,cada}, while later approaches enrich semantics by exploiting linguistic structure or stronger language models~\cite{gupta2021syntactically,zhu2024part,chen2024fine}. More recent works propose diffusion-powered alignment to mitigate the modality gap at the cost of heavier training and stochastic, noisy inference~\cite{tdsm}. Although effective, explicit alignment can remain fragile when label text only partially specifies the fine-grained dynamics that separate confusable actions, especially under challenging splits with substantial semantic ambiguity. In addition, generative ZSAR variants that model class-conditional feature distributions can reduce reliance on strict alignment, but they often lack an explicit, uncertainty-aware mechanism to discriminate among competing unseen classes during inference.

In this work, we propose SCALE, a lightweight and deterministic ZSAR framework that casts recognition as \emph{energy-based} comparison. We build a class-conditional variational autoencoder (CVAE) \cite{cVAE} in which frozen text representations parameterize both the latent prior and the decoder, enabling per-class likelihood evaluation for unseen actions without requiring unseen skeleton examples. Inspired by \cite{du2019implicit,zhai2016deep,al2022energy,lecun2006tutorial}, we define an energy function as the negative Evidence Lower Bound ($-$ELBO) and introduce a \emph{listwise} discrimination loss that aggregates negative classes with a semantic similarity bias to emphasize confusable negatives rather than treating all negatives uniformly. Crucially, we incorporate posterior uncertainty to adapt both (i) the decision margin and (ii) the loss contribution, inspired by uncertainty-aware and energy-based learning \cite{kendall2017uncertainties,charpentier2020posterior}, so that ambiguous samples are not over-penalized while confident samples enforce stricter separation. Finally, we add a latent prototype contrast term that aligns posterior means to text-derived latent prototypes, improving semantic organization of the latent space without requiring direct skeleton-text feature matching. Our contributions are summarized as follows:
\begin{itemize}
    \item We present SCALE, a class-conditional CVAE-based ZSAR framework that performs skeleton-based zero-shot recognition by ranking ELBO-derived energies over candidate unseen classes, avoiding iterative generation or test-time sampling.
    \item We introduce an uncertainty-aware, listwise energy discrimination objective that (i) emphasizes semantically similar hard negatives and (ii) leverages posterior uncertainty for adaptive margin relaxation and loss reweighting.
    \item We propose a latent prototype contrast loss that structures the latent space using text-derived prototypes, complementing ELBO-based modeling and improving class separability.
    \item We show our method's strong performance on NTU-60 and NTU-120 across multiple SynSE \cite{gupta2021syntactically} and PURLS \cite{zhu2024part} zero-shot splits, and provide ablations that validate the contribution of each component.
\end{itemize}

%
%
\section{Related Work}

\subsection{Zero-shot Skeleton-based Action Recognition}

Early ZSAR works learn shared pose–language spaces through metric or distributional alignment. \cite{jasani2019skeleton} combines ST-GCN \cite{stgcn} pose features with word embeddings and a relation module for unseen class compatibility. ReViSE \cite{hubert2017learning} uses modality-specific autoencoders to reduce distributional mismatch. CADA-VAE \cite{cada} employs modality-specific VAEs to create a shared latent space via distribution-level matching and cross-modal reconstruction. Within this VAE-based family, SynSE \cite{gupta2021syntactically} factorizes language by parts-of-speech and aligns structured linguistic factors with skeleton representations. MSF \cite{li2023multi} enriches semantics by combining action labels with dictionary-based and human-annotated descriptions. SA-DVAE \cite{li2024sa} disentangles skeleton latents into semantically relevant and nuisance components, aligning only the relevant subspace. SMIE \cite{zhou2023zero} maximizes mutual information between skeleton and text distributions while leveraging temporal evidence accumulation. To mitigate the semantic gap between high-level action labels and fine-grained motion, PURLS \cite{zhu2024part} uses large language models to generate body-part-aware descriptions and aligns them via cross-attention. STAR \cite{chen2024fine} decomposes skeletons into predefined body groups with learnable textual prompts. BSZSL \cite{liu2025beyond} augments inference with RGB semantic priors via multi-prompt guidance. More recently, generative inference has emerged as an alternative. TDSM \cite{tdsm} conditions a diffusion denoising process on text prompts and uses a triplet objective for discrimination.

Most existing ZSAR methods rely on explicit skeleton-text alignment, which can be fragile when language semantics only partially reflect fine-grained motion patterns. VAE-based approaches model class-conditional feature distributions but often lack explicit discrimination mechanisms for competing unseen classes, while diffusion-based methods improve robustness at the cost of heavy computational overhead and stochastic inference. In contrast, our method formulates ZSAR as a deterministic, listwise comparison of class-conditional energies from a lightweight CVAE, enabling effective discrimination under semantic ambiguity without iterative generation or test-time sampling.

\subsection{Energy-based and Uncertainty-aware Learning}

Energy-based learning formulates prediction and inference as energy function minimization. \cite{lecun2006tutorial} establishes energy-based learning as a unifying paradigm for classification, structured prediction, and ranking through margin-based, contrastive, and likelihood-driven objectives. Modern methods extend these principles to high-dimensional data and deep architectures. \cite{du2019implicit} shows that continuous energy-based models trained via maximum likelihood can scale to complex visual domains using learned energy functions rather than explicit generators. Similarly, \cite{zhai2016deep} parameterizes data likelihood with deterministic neural networks using score-matching to learn class-agnostic energy landscapes.

Energy-based formulations have been explored beyond closed-set classification. \cite{al2022energy} reinterprets classifier outputs as energy scores, showing that thresholding enables open-set recognition without prior access to unseen classes. In continual learning, \cite{li2022energy} formulates class prediction as a contrast over class-conditioned energies with a contrastive divergence objective maintaining separation between old and new classes.

Uncertainty-aware learning quantifies predictive uncertainty to improve robustness and calibration. \cite{kendall2017uncertainties} introduces a Bayesian framework jointly modeling aleatoric and epistemic uncertainty with uncertainty-dependent loss attenuation. \cite{charpentier2020posterior} models class-conditional evidence through latent-space density estimation. \cite{stutz2020confidence} enforces low-confidence predictions on adversarial inputs to avoid overconfident errors. Accordingly, we hypothesize that combining class-conditional energy modeling with uncertainty-aware margins and loss reweighting improves discriminability and robustness under semantic ambiguity.

%
%
\section{Methods}

\subsection{Problem Definition}

The task in zero-shot skeleton-based action recognition (ZSAR) is to recognize human actions whose classes are absent from the training skeleton data. Let $\mathcal{Y}^s$ and $\mathcal{Y}^u$ denote the sets of seen and unseen action classes, respectively, with $\mathcal{Y}^s \cap \mathcal{Y}^u = \varnothing$. The training set consists of skeleton sequences annotated only with seen labels,
\begin{equation}
\mathcal{D}_{\mathrm{train}} = \{(X_i, y_i)\}_{i=1}^{N}, \quad y_i \in \mathcal{Y}^s,
\end{equation}
where each skeleton sequence $X_i \in \mathbb{R}^{T \times V \times M \times C_{\mathrm{in}}}$ contains $T$ frames, $V$ joints, $M$ persons, and $C_{\mathrm{in}}$ input channels.

A pretrained skeleton encoder $f_s$ maps each sequence to a fixed-dimensional feature vector $x_s = f_s(X) \in \mathbb{R}^{D_s}$. Each action class $y$ is associated with a natural-language description $d_y$, encoded by a frozen pretrained text encoder $f_t$. The encoder produces two complementary representations: (1) a sequence of token-level embeddings $H_y \in \mathbb{R}^{L_y \times D_t}$, where each row corresponds to a token in the input text and $L_y$ is the number of tokens, capturing fine-grained semantic information at the word level; and (2) a global representation $h_y \in \mathbb{R}^{D_t}$, typically obtained via pooling over the token sequence, which summarizes the holistic semantic meaning of the entire text prompt. While $H_y$ preserves token-level structure and enables access to local semantic details, $h_y$ provides a compact, sequence-level semantic anchor. These text embeddings are available for both seen and unseen classes and serve as the only source of supervision for unseen actions.

\subsection{Overview of our Proposed Framework}

\begin{figure}[tbp]
  \centering
  \includegraphics[width=1.0\textwidth]{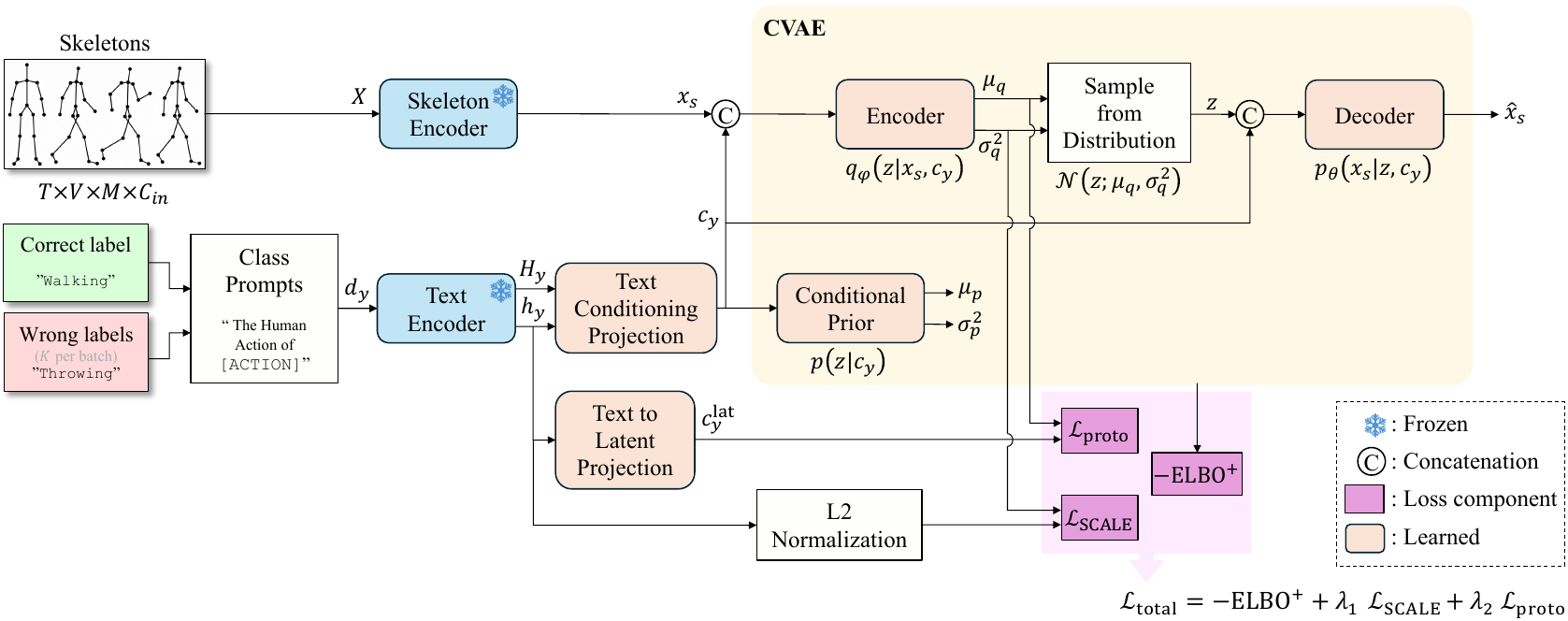}
  \vspace{-6mm}
  \caption{Training pipeline of our proposed SCALE framework.}
  \label{fig:train_framework}
  \vspace{-1mm}
\end{figure}

Our framework, shown in Fig.~\ref{fig:train_framework}, builds upon a conditional variational autoencoder (CVAE) \cite{cVAE} that models the distribution of skeleton features conditioned on class-level semantics. Rather than directly aligning skeleton and text features, we learn a shared latent representation where text descriptions parameterize class-conditional priors and latent prototypes, while skeleton features are mapped through conditional encoding. The latent space is jointly shaped by generative consistency and semantic discrimination.

To enhance discriminability and robustness, we augment the generative objective with: (1) an energy-based loss contrasting positive and negative class hypotheses in ELBO space, and (2) a latent prototype contrast loss structuring the latent space according to text-derived class prototypes.

\subsection{Skeleton and Text Feature Encoders}
\label{sec43}

We adopt the pretrained Shift-GCN~\cite{cheng2020skeleton} model as the skeleton encoder backbone. Given an input skeleton sequence $X \in \mathbb{R}^{T \times V \times M \times C_{\mathrm{in}}}$, the skeleton encoder $f_s$ generates a feature vector $x_s = f_s(X) \in \mathbb{R}^{D_s}$. The encoder is pretrained on seen data using supervised classification objective and then frozen, serving purely as a feature extractor.

For text encoding, we use class labels as textual input with a simple prompt $d_y=\{\text{The Human Action of [ACTION]}\}$, processed by the pretrained text encoder of CLIP~\cite{clip}, $f_t$, to ensure fair comparison with existing methods \cite{zhu2024part,chen2024fine,li2024sa}. The encoder produces both token-level embeddings $H_y$ and the global representation $h_y$, which is obtained by End-of-Text (EOT) pooling in CLIP's architecture~\cite{clip}, as described in the problem definition.

\subsection{Text Conditioning from Language Descriptions}
\label{sec44}

Unlike approaches that directly align skeleton and text features~\cite{zhu2024part,li2024sa}, we use language information exclusively to parameterize class-conditional priors and latent prototypes. To bridge the gap between text feature space and latent variables, we introduce a learnable text projection module $f_c$ that transforms text representations into a compact conditioning vector:
\begin{equation}
    c_y = f_c(h_y, H_y) = \mathrm{Concat}\!\left(\mathrm{MLP}(h_y),\; \mathrm{MLP}(\mathrm{Pool}(H_y))\right) \in \mathbb{R}^{D_c},
\end{equation}
where $f_c$ applies separate Multilayer Perceptrons (MLPs) to $h_y$ and average pooled $H_y$, then concatenates and transforms the results.

Given $c_y$, a conditional prior network predicts the parameters of a Gaussian distribution over the latent variable $z$:
\begin{equation}
p_\theta(z \mid c_y) = \mathcal{N}\big(z; \mu_p(c_y), \sigma_p^2(c_y) I\big).
\end{equation}
This prior defines a class-specific region in latent space and can be computed for unseen classes without requiring skeletons.

\subsection{CVAE Framework}
\label{sec45}

We utilize a CVAE to model the distribution of skeleton features. The inference network (encoder) approximates the posterior distribution over latent variables $z$ given the input skeleton feature $x_s$ and the class-specific condition $c_y$:
\begin{equation}
q_\phi(z \mid x_s, c_y) = \mathcal{N}\big(z; \mu_q(x_s, c_y), \sigma_q^2(x_s, c_y) I\big),
\end{equation}
where the encoder is an MLP operating on the concatenation $[x_s; c_y]$. The generative network (decoder) reconstructs skeleton features:
\begin{equation}
p_\theta(x_s \mid z, c_y) = \mathcal{N}\big(x_s; \mu_d(z, c_y), \sigma_d^2(z, c_y) I\big),
\end{equation}
where a latent sample $z$ is drawn from the posterior.

\subsection{Training Objectives}
\label{sec46}

We optimize a composite objective combining generative reconstruction, energy-based discrimination, and latent prototype alignment. For a skeleton feature $x_s$ and class condition $c_y$, the ELBO is:
\begin{equation}
\mathrm{ELBO}(x_s, c_y)
=
\mathbb{E}_{q_\phi(z \mid x_s, c_y)}
\big[\log p_\theta(x_s \mid z, c_y)\big]
-
\beta \, D_{\mathrm{KL}}\big(q_\phi(z \mid x_s, c_y) \,\|\, p_\theta(z \mid c_y)\big),
\end{equation}
where the first term encourages accurate reconstruction of skeleton features and the second term regularizes the approximate posterior toward the class-conditional prior. The hyperparameter $\beta$ mitigates posterior collapse \cite{higgins2017beta}. We minimize $-\mathrm{ELBO}^+$ for positive skeleton-class pairs.

\subsubsection{Energy-based Discrimination.}

While the ELBO measures how well a skeleton feature is explained by a class-conditional model, it does not explicitly enforce separation between competing hypotheses. This becomes problematic in ZSAR, where multiple actions may exhibit similar skeletal dynamics and differ only in subtle motion cues. We, therefore, introduce an energy-based discrimination objective that contrasts compatibility with the correct class against negative classes.

We define energy as $E(x_s, c_y) = -\mathrm{ELBO}(x_s, c_y)$, so lower energy indicates higher compatibility. For a sample with positive class $y^+$ and $K$ negative classes $\{y_j^-\}_{j=1}^{K}$ within a batch, we aggregate negative energies using a similarity-biased soft maximum. Using normalized text embeddings $\hat h_y = h_y / \|h_y\|_2$, we define semantic similarity $s_j = \hat h_{y^+}^\top \hat h_{y_j^-}$ and compute:
\begin{equation}
\tilde{E}_{\mathrm{neg}}(x_s)
=
\log \sum_{j=1}^{K} \exp\big(E(x_s, c_{y_j^-}) + \alpha s_j\big),
\end{equation}
where $\alpha>0$ controls semantic bias strength and increases the contribution of semantically similar negatives.

We incorporate posterior uncertainty to adapt both margin and loss weight. For the diagonal Gaussian posterior $q_\phi(z \mid x_s, c_{y^+}) = \mathcal{N}(z;\mu_q,\mathrm{diag}(\sigma_q^2))$, we define uncertainty as the average posterior variance over all latent dimensions:
\begin{equation}
u(x_s, c_{y^+})=\frac{1}{D_z}\sum_{d=1}^{D_z}\sigma_{q,d}^2.
\end{equation}
The energy gap is then:
\begin{equation}
\Delta(x_s)=E(x_s, c_{y^+})-\tilde{E}_{\mathrm{neg}}(x_s)+\tau-\beta_u \, u(x_s,c_{y^+}),
\end{equation}
where $\tau>0$ is a base margin and $\beta_u>0$ controls uncertainty-adaptive margin relaxation.

We compute the \emph{Semantic- and Confidence-Aware Listwise Energy (SCALE)} loss as:
\begin{equation}
\mathcal{L}_{\mathrm{SCALE}}
=
\exp\big(-\gamma \, u(x_s, c_{y^+})\big) \cdot
\log\big(1 + \exp(\Delta(x_s))\big),
\end{equation}
where $\gamma>0$ controls uncertainty-based reweighting. The factor $\exp\!\big(-\gamma\,u(x_s, c_{y^+})\big)$ downweights high-uncertainty samples, while the margin $\tau-\beta_u\,u(x_s, c_{y^+})$ relaxes separation requirements for uncertain samples and enforces stricter margins for confident ones.

\subsubsection{Latent Prototype Contrast.}

We impose semantic structure directly in latent space by treating the posterior mean $\mu_q(x_s, c_{y^+}) \in \mathbb{R}^{D_z}$ as the latent representation. For each action class $y$, we define a latent prototype $c_y^{\mathrm{lat}} = g(h_y) \in \mathbb{R}^{D_z}$ via a learnable MLP projection $g(\cdot)$. The contrastive loss is:
\begin{equation}
\mathcal{L}_{\mathrm{proto}}(x_s,y^+)
=
-\log
\frac{\exp\!\big(\mathrm{sim}(\mu_q(x_s, c_{y^+}), c_{y^+}^{\mathrm{lat}})/\lambda\big)}
{\sum_j \exp\!\big(\mathrm{sim}(\mu_q(x_s, c_{y^+}), c_{j}^{\mathrm{lat}})/\lambda\big)},
\end{equation}
where $\mathrm{sim}(\cdot,\cdot)$ is cosine similarity and $\lambda>0$ is a temperature parameter. This complements ELBO-based regularization and encourages semantically organized latent space.

\subsubsection{Overall Training Loss.}

The overall training objective is:
\begin{equation}
\mathcal{L}_{\mathrm{total}} =
- \mathrm{ELBO}^+
+
\lambda_1\cdot \mathcal{L}_{\mathrm{SCALE}}
+
\lambda_2\cdot \mathcal{L}_{\mathrm{proto}},
\end{equation}
where $\lambda_1>0$ and $\lambda_2>0$ are hyperparameters that control the effect of the proposed loss components.

\subsection{Inference}
\label{sec47}

\begin{figure}[tbp]
  \centering
  \includegraphics[width=1.0\textwidth]{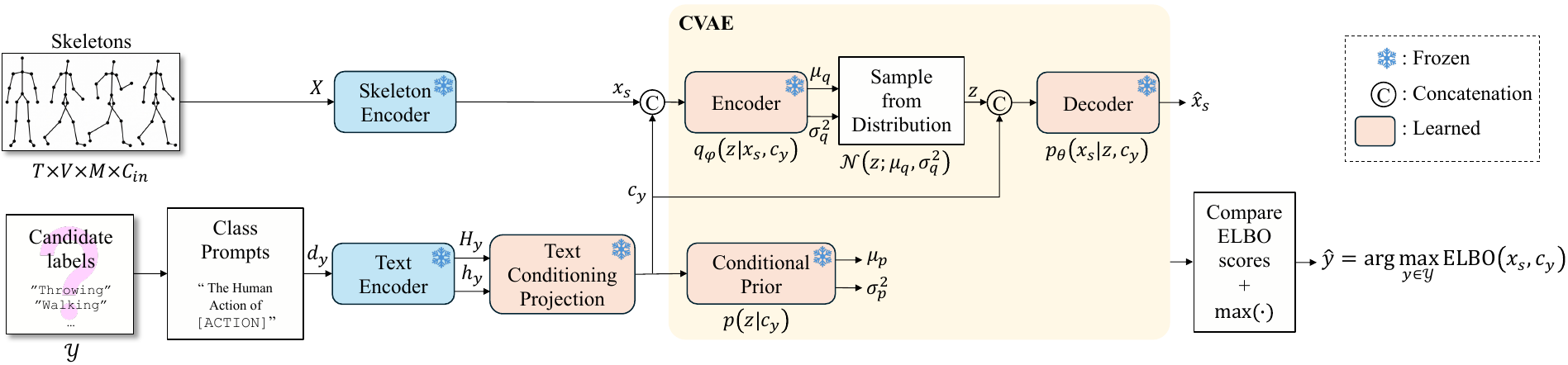}
  \vspace{-6mm}
  \caption{Inference pipeline of our proposed SCALE framework.}
  \label{fig:inference_framework}
  \vspace{-1mm}
\end{figure}

As shown in Fig.~\ref{fig:inference_framework}, at inference time, given a test skeleton sequence $X$, we extract its feature $x_s = f_s(X)$ and consider candidate unseen classes $\mathcal{Y}^{u}$ with text descriptions $\{d_y\}$. For each candidate class $y \in \mathcal{Y}^{u}$, we process the text description through the frozen text encoder and map it to conditioning vector $c_y$ via $f_c$. We then evaluate $\mathrm{ELBO}(x_s, c_y)$ by computing the approximate posterior $q_\phi(z \mid x_s, c_y)$ and corresponding reconstruction and regularization terms, and predict the label as:
\begin{equation}
\hat{y}
=
\arg\max_{y \in \mathcal{Y}^{u}}
\mathrm{ELBO}(x_s, c_y).
\end{equation}

%
%
\section{Experiments}

\subsection{Datasets}

\subsubsection{NTU RGB+D \cite{ntu}.}
The NTU-60 dataset consists of 56,880 skeleton sequences spanning 60 action classes, captured by depth sensors from 40 subjects under multiple camera viewpoints. Following benchmarks in \cite{gupta2021syntactically,zhu2024part}, we evaluate SCALE under four [seen/unseen] splits: [55/5] and [48/12] in SynSE (mild zero-shot settings), and [40/20] and [30/30] in PURLS (more challenging scenarios).

\subsubsection{NTU RGB+D 120 \cite{ntu120}.}
NTU-120 extends NTU-60 to 120 action categories with 114,480 sequences from 106 subjects across 155 viewpoints. We evaluate under four zero-shot splits: [110/10] and [96/24] in SynSE, and [80/40] and [60/60] in PURLS \cite{gupta2021syntactically,zhu2024part}. For all splits, skeleton sequences from unseen classes are excluded during training, with textual descriptions serving as the only supervision for unseen classes at inference.

\subsection{Experimental Setup}

We implement SCALE using PyTorch \cite{pytorch} on a single NVIDIA RTX~4090 GPU. For skeleton feature extraction, we adopt Shift-GCN~\cite{cheng2020skeleton} as the skeleton encoder, pretrained on seen data using cross-entropy and then frozen. Features are extracted from the layer preceding the classification head. For textual encoding, we employ the frozen CLIP \cite{clip} text encoder (ViT-B/32) with simple prompts based on class labels for fair comparison.

We train our model using the AdamW \cite{adamW} optimizer with learning rate $10^{-4}$ and weight decay $10^{-3}$, with cosine learning rate scheduling \cite{cosineSch}. Batch size is 256 for both training and testing. We adopt $\beta$ cyclical annealing \cite{cyclical} to mitigate posterior collapse, using four cycles, each linearly increasing $\beta$ from 0 to 1.0.

The following hyperparameters are fixed across all experiments: semantic similarity weight $\alpha=1.0$, uncertainty-adaptive margin coefficient $\beta_u=0.5$, base margin $\tau=1.0$, uncertainty reweighting factor $\gamma=2.0$, temperature parameter $\lambda=0.2$, and loss weights $\lambda_1=1.0$ and $\lambda_2=0.5$. All MLP projection modules consist of a single hidden layer whose dimensionality is set to $1/4$ of the input feature size. Concretely, the hidden dimension is set to 64 for skeleton feature projections, corresponding to the 256-dimensional skeleton feature vectors, and to 128 for text feature projections, corresponding to the 512-dimensional CLIP text embeddings. All hyperparameters are empirically selected via grid search.

\subsection{Comparison with State-of-the-Art Methods}

Table~\ref{tab:results} reports top-1 accuracy comparisons with recent ZSAR methods on NTU-60 and NTU-120 under multiple [seen/unseen] splits, covering early embedding-based approaches, VAE-based generative models, and alignment-based frameworks.

\begin{table}[tp]
    \scriptsize
    \centering
    \resizebox{1.0\textwidth}{!}{
    \def\arraystretch{1.3}
    \begin{tabular} {c|c|c|c|c|c|c|c|c|c}
        \Xhline{2\arrayrulewidth}
        \multirow{2}{*}{Method} & \multirow{2}{*}{Venue} &\multicolumn{4}{c|}{NTU-60 (Acc, \%)} & \multicolumn{4}{c}{NTU-120 (Acc, \%)} \\
        \cline{3-10}
        & & [55/5] & [48/12] & [40/20] & [30/30] & [110/10] & [96/24] & [80/40] & [60/60] \\
        \Xhline{2\arrayrulewidth}
        ReViSE \cite{hubert2017learning} & ICCV2017 & 53.9 & 17.5 & 24.3 & 14.8 & 55.0 & 32.4 & 19.5 & 8.3 \\
        JPoSE \cite{wray2019fine} & ICCV2019 & 64.8 & 28.8 & 20.1 & 12.4 & 51.9 & 32.4 & 13.7 & 7.7 \\
        CADA-VAE \cite{cada} & CVPR2019 & 76.8 & 29.0 & 16.2 & 11.5 & 59.5 & 35.8 & 10.6 & 5.7 \\
        SynSE \cite{gupta2021syntactically} & ICIP2021 & 75.8 & 33.3 & 19.9 & 12.0 & 62.7 & 38.7 & 13.6 & 7.7 \\
        SMIE \cite{zhou2023zero} & ACMM2023 & 78.0 & 40.2 & - & - & 65.7 & 45.3 & - & - \\
        SA-DVAE \cite{li2024sa} & ECCV2024 & \textcolor{ForestGreen}{82.4} & 41.4 & - & - & 68.8 & 46.1 & - & - \\
        STAR \cite{chen2024fine} & ACMM2024 & 81.4 & \textcolor{ForestGreen}{45.1} & - & - & 63.3 & 44.3 & - & - \\
        PURLS \cite{zhu2024part} & CVPR2024 & 79.2 & 41.0 & \textcolor{ForestGreen}{31.0} & \textcolor{ForestGreen}{23.5} & \textcolor{ForestGreen}{72.0} & \textcolor{ForestGreen}{52.0} & \textcolor{ForestGreen}{28.4} & \textcolor{ForestGreen}{19.6} \\
        \Xhline{2\arrayrulewidth}
        TDSM \cite{tdsm} & ICCV2025 & \textcolor{red}{86.5} & \textcolor{red}{56.0} & \textcolor{red}{36.1} & \textcolor{blue}{25.9} & \textcolor{red}{74.2} & \textcolor{red}{65.1} & \textcolor{red}{37.0} & \textcolor{red}{27.2} \\
        \textbf{SCALE (Ours)} & - & \textbf{\textcolor{blue}{84.5}}
 & \textbf{\textcolor{blue}{50.0}} & \textbf{\textcolor{blue}{35.5}} & \textbf{\textcolor{red}{27.7}} & \textbf{\textcolor{blue}{73.6}} & \textbf{\textcolor{blue}{54.9}} & \textbf{\textcolor{blue}{32.1}} & \textbf{\textcolor{blue}{23.5}}
    \end{tabular}}
    \vspace{0cm}
    \caption{Top-1 accuracy (\%) comparison of ZSAR methods on the NTU-60 and NTU-120 datasets under multiple [seen/unseen] class splits under SynSE \cite{gupta2021syntactically} and PURLS \cite{zhu2024part} benchmarks. Results in \textcolor{red}{red} denote the best performance, while those in \textcolor{blue}{blue} indicate the second best and those in \textcolor{ForestGreen}{green} indicate the third best.}
  \label{tab:results}
  \vspace{-4mm}
\end{table}

SCALE achieves the second-best overall performance, consistently outperforming all methods except TDSM~\cite{tdsm} across both datasets and benchmarks. Compared to the third-best methods, SCALE achieves performance gains of $2.1\%$, $4.9\%$, $4.5\%$, and $4.2\%$ on NTU-60 splits [55/5], [48/12], [40/20], and [30/30], respectively, and $1.6\%$, $2.9\%$, $3.7\%$, and $3.9\%$ on NTU-120 splits [110/10], [96/24], [80/40], and [60/60], respectively. The performance gap widens under challenging splits with higher unseen class proportions, indicating that as semantic ambiguity increases, direct skeleton-text alignment becomes less effective, while our class-conditional energy-based approach provides more robust discrimination.
\vspace{-3mm}

\subsubsection{Comparison with TDSM.}
TDSM~\cite{tdsm} outperforms SCALE on 7 of 8 benchmarks, with SCALE achieving superior performance on NTU-60 [30/30]. However, TDSM's gains come at substantial computational cost, as quantified in Table~\ref{tab:tdsm_computation}. TDSM employs diffusion transformers~\cite{DiT} with $261.21$M parameters ($\sim$107$\times$ more than SCALE's $2.43$M) and requires 10 independent inference trials with averaging, multiplying effective inference time to $103.14$ ms per sample compared to SCALE's $0.454$ ms ($\sim$227$\times$ faster). Similarly, TDSM's effective computational cost is $32.03$ GFLOPs (10 trials) compared to SCALE's $0.002$ GFLOPs ($\sim$16,000$\times$ more efficient). In contrast, SCALE performs deterministic single-pass inference via direct ELBO ranking, achieving within $2.0\%$ of TDSM on 3 of 8 benchmarks with dramatically lower computational overhead, demonstrating superior efficiency-accuracy trade-offs.

\begin{table}[tp]
    \scriptsize
    \centering
    \resizebox{\textwidth}{!}{
    \def\arraystretch{1.3}
    \begin{tabular}{l!{\vrule width 1pt}c!{\vrule width 1pt}c|c!{\vrule width 1pt}c|c}
        \Xhline{2\arrayrulewidth}
        Method 
        & Parameters (M) 
        & GFLOPs 
        & Effective GFLOPs 
        & Inference Time (ms) 
        & Effective Inference Time (ms) \\
        \Xhline{2\arrayrulewidth}
        TDSM~\cite{tdsm} 
        & 261.21 
        & 3.203
        & 32.03$^{\dagger}$
        & 10.314
        & 103.14$^{\dagger}$ \\
        \textbf{SCALE (Ours)} 
        & \textbf{2.43} 
        & \textbf{0.002} 
        & \textbf{0.002}
        & \textbf{0.454} 
        & \textbf{0.454} \\
        \Xhline{2\arrayrulewidth}
        \addlinespace[0.5mm]
        \multicolumn{6}{l}{\footnotesize $^\dagger$ TDSM averages over 10 inference trials.} \\
    \end{tabular}}
    \vspace{0.5mm}
    \caption{Computational efficiency comparison between SCALE and TDSM. Inference time and GFLOPs are measured per sample.}
    \label{tab:tdsm_computation}
\end{table}

\subsection{Ablation Studies}

\subsubsection{Effect of Loss Components.}

Table~\ref{tab:ablation_loss} shows the contribution of each loss component by progressively adding $\mathcal{L}_{\mathrm{proto}}$ and $\mathcal{L}_{\mathrm{SCALE}}$ to the CVAE baseline. The $-$ELBO$^+$ baseline yields the weakest performance ($78.7\%$ on NTU-60 [55/5], $65.1\%$ on NTU-120 [110/10]). Adding $\mathcal{L}_{\mathrm{proto}}$ improves performance by $+2.4\%$ and $+3.9\%$ on these splits, while adding $\mathcal{L}_{\mathrm{SCALE}}$ yields larger gains of $+4.0\%$ and $+5.5\%$, indicating complementary benefits. The full SCALE loss achieves the best results ($84.5\%$ and $73.6\%$), with improvements of $+5.8\%$ and $+8.5\%$ over the baseline, demonstrating that jointly enforcing semantic structure in latent space and uncertainty-aware listwise discrimination in ELBO space is crucial for robust ZSAR.

\begin{table}[tp]
    \scriptsize
    \centering
    \resizebox{0.8\textwidth}{!}{
    \def\arraystretch{1.3}
    \begin{tabular}{l|c|c|c|c}
        \Xhline{2\arrayrulewidth}
        \multirow{2}{*}{Training Loss} 
        & \multicolumn{2}{c|}{NTU-60 (Acc, \%)} 
        & \multicolumn{2}{c}{NTU-120 (Acc, \%)} \\
        \cline{2-5}
        & [55/5] & [48/12] & [110/10] & [96/24] \\
        \Xhline{2\arrayrulewidth}
        $-\mathrm{ELBO}^+$ 
        & 78.7 & 43.3 & 65.1 & 47.1 \\
        $-\mathrm{ELBO}^+ + \mathcal{L}_{\mathrm{proto}}$ 
        & 81.1$^{\textcolor{ForestGreen}{\uparrow 2.4}}$ & 45.7$^{\textcolor{ForestGreen}{\uparrow 2.4}}$ & 69.0$^{\textcolor{ForestGreen}{\uparrow 3.9}}$ & 50.6$^{\textcolor{ForestGreen}{\uparrow 3.5}}$ \\
        $-\mathrm{ELBO}^+ + \mathcal{L}_{\mathrm{SCALE}}$ 
        & 82.7$^{\textcolor{ForestGreen}{\uparrow 4.0}}$ & 46.6$^{\textcolor{ForestGreen}{\uparrow 3.3}}$ & 70.6$^{\textcolor{ForestGreen}{\uparrow 5.5}}$ & 52.2$^{\textcolor{ForestGreen}{\uparrow 5.1}}$ \\
        \textbf{$-\mathrm{ELBO}^+ + \mathcal{L}_{\mathrm{SCALE}} + \mathcal{L}_{\mathrm{proto}}$} 
        & \textbf{84.5}$^{\textcolor{ForestGreen}{\uparrow 5.8}}$ & \textbf{50.0}$^{\textcolor{ForestGreen}{\uparrow 6.7}}$ & \textbf{73.6}$^{\textcolor{ForestGreen}{\uparrow 8.5}}$ & \textbf{54.9}$^{\textcolor{ForestGreen}{\uparrow 7.8}}$ \\
        \Xhline{2\arrayrulewidth}
    \end{tabular}}
    \vspace{1.5mm}
    \caption{Ablation study on the effect of different loss components. Top-1 accuracy (\%) is reported on NTU-60 and NTU-120 under SynSE \cite{gupta2021syntactically} settings.}
    \label{tab:ablation_loss}
    \vspace{-3mm}
\end{table}

\subsubsection{Effect of Semantic Similarity Bias.}

Table~\ref{tab:ablation_alpha} shows the effect of semantic similarity bias parameter $\alpha$. When $\alpha=0$ (equal negative contribution), performance drops to $83.9\%$ on NTU-60 [55/5] and $72.9\%$ on NTU-120 [110/10]. Increasing $\alpha$ to $1.0$ improves performance by $+0.6\%$ and $+0.7\%$, confirming that prioritizing semantically confusable negatives is beneficial. Further increases to $\alpha=2.0$ or $3.0$ yield marginal gains or slight degradation, suggesting that overly large $\alpha$ overemphasizes a small subset of negatives and reduces generalization. The optimal $\alpha=1.0$ provides the best balance between hard negative emphasis and stable optimization.

\begin{table}[tp]
    \scriptsize
    \centering
    \resizebox{0.5\textwidth}{!}{
    \def\arraystretch{1.3}
    \begin{tabular}{l|c|c|c|c}
        \Xhline{2\arrayrulewidth}
        \multirow{2}{*}{$\alpha$} 
        & \multicolumn{2}{c|}{NTU-60 (Acc, \%)} 
        & \multicolumn{2}{c}{NTU-120 (Acc, \%)} \\
        \cline{2-5}
        & [55/5] & [48/12] & [110/10] & [96/24] \\
        \Xhline{2\arrayrulewidth}
        $\alpha = 0$ 
        & 83.9 & 49.6 & 72.9 & 54.3 \\
        $\boldsymbol{\alpha} = \mathbf{1.0}$ 
        & \textbf{\textcolor{red}{84.5}}$^{\textcolor{ForestGreen}{\uparrow 0.6}}$ & \textbf{50.0}$^{\textcolor{ForestGreen}{\uparrow 0.4}}$ & \textbf{\textcolor{red}{73.6}}$^{\textcolor{ForestGreen}{\uparrow 0.7}}$ & \textbf{\textcolor{red}{54.9}}$^{\textcolor{ForestGreen}{\uparrow 0.6}}$ \\
        $\alpha = 2.0$ 
        & 84.3$^{\textcolor{ForestGreen}{\uparrow 0.4}}$ & \textcolor{red}{50.1}$^{\textcolor{ForestGreen}{\uparrow 0.5}}$ & 73.5$^{\textcolor{ForestGreen}{\uparrow 0.6}}$ & 54.4$^{\textcolor{ForestGreen}{\uparrow 0.1}}$ \\
        $\alpha = 3.0$ 
        & 84.4$^{\textcolor{ForestGreen}{\uparrow 0.5}}$ & 49.8$^{\textcolor{ForestGreen}{\uparrow 0.2}}$ & 73.3$^{\textcolor{ForestGreen}{\uparrow 0.4}}$ & 54.5$^{\textcolor{ForestGreen}{\uparrow 0.2}}$ \\
        \Xhline{2\arrayrulewidth}
    \end{tabular}}
    \vspace{1.5mm}
    \caption{Ablation study on the semantic similarity bias parameter $\alpha$ in the SCALE loss. Top-1 accuracy (\%) is reported on NTU-60 and NTU-120 under SynSE \cite{gupta2021syntactically} settings. Results in \textcolor{red}{red} denote the best performance.}
    \label{tab:ablation_alpha}
    \vspace{-3mm}
\end{table}

\subsubsection{Effect of Uncertainty Modeling $u(x_s, c_y)$.}
Table~\ref{tab:ablation_uncertainty} analyzes uncertainty modeling contributions by isolating margin relaxation ($\beta_u$) and loss reweighting ($\gamma$). The baseline (fixed margin, no reweighting) achieves $83.7\%$ on NTU-60 [55/5] and $73.1\%$ on NTU-120 [110/10]. Uncertainty-adaptive margin only ($\beta_u=0.5,\gamma=0$) improves by $+0.6\%$ and $+0.4\%$, with larger gains on challenging splits ($+1.2\%$ and $+1.3\%$ on [48/12] and [96/24]), indicating that relaxing margins for uncertain samples prevents over-penalization under high class ambiguity. Uncertainty-based reweighting only ($\beta_u=0,\gamma=2.0$) yields smaller improvements ($+0.3\%$ and $+0.2\%$), reducing the influence of noisy samples. The full formulation ($\beta_u=0.5,\gamma=2.0$) produces the largest improvements ($+0.8\%$ and $+0.5\%$, up to $+1.8\%$ on [48/12]), demonstrating that both mechanisms are complementary and essential.

\begin{table}[tbp]
    \scriptsize
    \centering
    \resizebox{0.85\textwidth}{!}{
    \def\arraystretch{1.3}
    \begin{tabular}{l|c|c|c|c|c|c}
        \Xhline{2\arrayrulewidth}
        \multirow{2}{*}{Uncertainty Modeling} 
        & \multirow{2}{*}{$\beta_u$} 
        & \multirow{2}{*}{$\gamma$} 
        & \multicolumn{2}{c|}{NTU-60 (Acc, \%)} 
        & \multicolumn{2}{c}{NTU-120 (Acc, \%)} \\
        \cline{4-7}
        & & & [55/5] & [48/12] & [110/10] & [96/24] \\
        \Xhline{2\arrayrulewidth}

        Fixed margin, No reweighting 
        & 0 & 0 
        & 83.7 & 48.2 & 73.1 & 53.2 \\

        Uncertainty-adaptive margin only 
        & 0.5 & 0 
        & 84.3$^{\textcolor{ForestGreen}{\uparrow 0.6}}$
        & 49.4$^{\textcolor{ForestGreen}{\uparrow 1.2}}$
        & 73.5$^{\textcolor{ForestGreen}{\uparrow 0.4}}$
        & 54.5$^{\textcolor{ForestGreen}{\uparrow 1.3}}$ \\

        Uncertainty-based reweighting only 
        & 0 & 2.0 
        & 84.0$^{\textcolor{ForestGreen}{\uparrow 0.3}}$
        & 48.6$^{\textcolor{ForestGreen}{\uparrow 0.4}}$
        & 73.3$^{\textcolor{ForestGreen}{\uparrow 0.2}}$
        & 53.9$^{\textcolor{ForestGreen}{\uparrow 0.7}}$ \\

        \textbf{Full $\boldsymbol{\mathcal{L}_{\mathrm{SCALE}}}$ loss} 
        & \textbf{0.5} & \textbf{2.0} 
        & \textbf{84.5}$^{\textcolor{ForestGreen}{\uparrow 0.8}}$
        & \textbf{50.0}$^{\textcolor{ForestGreen}{\uparrow 1.8}}$
        & \textbf{73.6}$^{\textcolor{ForestGreen}{\uparrow 0.5}}$
        & \textbf{54.9}$^{\textcolor{ForestGreen}{\uparrow 1.7}}$ \\

        \Xhline{2\arrayrulewidth}
    \end{tabular}}
    \vspace{1.5mm}
    \caption{Ablation study on uncertainty modeling in the SCALE loss. We analyze the individual and joint effects of uncertainty-adaptive margin relaxation (controlled by $\beta_u$) and uncertainty-based loss reweighting (controlled by $\gamma$). Results are shown on NTU-60 and NTU-120 under SynSE~\cite{gupta2021syntactically} splits.}
    \label{tab:ablation_uncertainty}
    \vspace{-4mm}
\end{table}

%
%
\section{Conclusion}

In this paper, we present SCALE, an uncertainty-aware energy-based framework for zero-shot skeleton-based action recognition that avoids explicit skeleton-text alignment. By modeling class-conditional skeleton feature distributions with a lightweight CVAE and casting recognition as a deterministic, listwise comparison of ELBO-derived energies, SCALE enables effective discrimination among unseen action classes using only textual descriptions. The proposed uncertainty-adaptive margin and loss reweighting leverage posterior variance to mitigate semantic ambiguity and noisy supervision, while latent prototype contrast further organizes the latent space according to class semantics. Extensive experiments on NTU-60 and NTU-120 under multiple zero-shot splits demonstrate that SCALE achieves the second-best overall performance, consistently outperforming prior VAE-based and alignment-based methods by substantial margins (e.g., $+4.8\%$ over SA-DVAE on NTU-120 [110/10], $+4.5\%$ over PURLS on NTU-60 [40/20]), and remains highly competitive with computationally intensive diffusion-based approaches while offering deterministic single-pass inference and superior efficiency-accuracy trade-offs.

\subsubsection{Acknowledgements} This paper is supported by the Natural Sciences and Engineering Research Council of Canada (NSERC) under grant RGPIN-2020-04525.

%
%
\bibliographystyle{splncs04}
\bibliography{citations}

@inproceedings{clip,
  title={Learning transferable visual models from natural language supervision},
  author={Radford, Alec and Kim, Jong Wook and Hallacy, Chris and Ramesh, Aditya and Goh, Gabriel and Agarwal, Sandhini and Sastry, Girish and Askell, Amanda and Mishkin, Pamela and Clark, Jack and others},
  booktitle={International conference on machine learning},
  pages={8748--8763},
  year={2021},
  organization={PmLR}
}

@inproceedings{chen2024fine,
  title={Fine-grained side information guided dual-prompts for zero-shot skeleton action recognition},
  author={Chen, Yang and Guo, Jingcai and He, Tian and Lu, Xiaocheng and Wang, Ling},
  booktitle={Proceedings of the 32nd ACM International Conference on Multimedia},
  pages={778--786},
  year={2024}
}

@inproceedings{cheng2020skeleton,
  title={Skeleton-based action recognition with shift graph convolutional network},
  author={Cheng, Ke and Zhang, Yifan and He, Xiangyu and Chen, Weihan and Cheng, Jian and Lu, Hanqing},
  booktitle={Proceedings of the IEEE/CVF conference on computer vision and pattern recognition},
  pages={183--192},
  year={2020}
}

@inproceedings{li2024sa,
  title={Sa-dvae: Improving zero-shot skeleton-based action recognition by disentangled variational autoencoders},
  author={Li, Sheng-Wei and Wei, Zi-Xiang and Chen, Wei-Jie and Yu, Yi-Hsin and Yang, Chih-Yuan and Hsu, Jane Yung-jen},
  booktitle={European Conference on Computer Vision},
  pages={447--462},
  year={2024},
  organization={Springer}
}

@inproceedings{zhu2024part,
  title={Part-aware unified representation of language and skeleton for zero-shot action recognition},
  author={Zhu, Anqi and Ke, Qiuhong and Gong, Mingming and Bailey, James},
  booktitle={Proceedings of the IEEE/CVF Conference on Computer Vision and Pattern Recognition},
  pages={18761--18770},
  year={2024}
}

@article{adamW,
  title={Decoupled weight decay regularization},
  author={Loshchilov, Ilya and Hutter, Frank},
  journal={arXiv preprint arXiv:1711.05101},
  year={2017}
}

@article{cosineSch,
  title={Sgdr: Stochastic gradient descent with warm restarts},
  author={Loshchilov, Ilya and Hutter, Frank},
  journal={arXiv preprint arXiv:1608.03983},
  year={2016}
}

@inproceedings{oraki2024lortsar,
  title={LORTSAR: Low-Rank Transformer for Skeleton-Based Action Recognition},
  author={Oraki, Soroush and Zhuang, Harry and Liang, Jie},
  booktitle={International Symposium on Visual Computing},
  pages={196--207},
  year={2024},
  organization={Springer}
}

@article{cVAE,
  title={Learning structured output representation using deep conditional generative models},
  author={Sohn, Kihyuk and Lee, Honglak and Yan, Xinchen},
  journal={Advances in neural information processing systems},
  volume={28},
  year={2015}
}

@inproceedings{higgins2017beta,
  title={beta-vae: Learning basic visual concepts with a constrained variational framework},
  author={Higgins, Irina and Matthey, Loic and Pal, Arka and Burgess, Christopher and Glorot, Xavier and Botvinick, Matthew and Mohamed, Shakir and Lerchner, Alexander},
  booktitle={International conference on learning representations},
  year={2017}
}

@article{jasani2019skeleton,
  title={Skeleton based zero shot action recognition in joint pose-language semantic space},
  author={Jasani, Bhavan and Mazagonwalla, Afshaan},
  journal={arXiv preprint arXiv:1911.11344},
  year={2019}
}

@InProceedings{tdsm,
    author    = {Do, Jeonghyeok and Kim, Munchurl},
    title     = {Bridging the Skeleton-Text Modality Gap: Diffusion-Powered Modality Alignment for Zero-shot Skeleton-based Action Recognition},
    booktitle = {Proceedings of the IEEE/CVF International Conference on Computer Vision (ICCV)},
    month     = {October},
    year      = {2025},
    pages     = {12757-12768}
}

@article{cyclical,
  title={Cyclical annealing schedule: A simple approach to mitigating kl vanishing},
  author={Fu, Hao and Li, Chunyuan and Liu, Xiaodong and Gao, Jianfeng and Celikyilmaz, Asli and Carin, Lawrence},
  journal={arXiv preprint arXiv:1903.10145},
  year={2019}
}

@article{ding2025lstc,
  title={LSTC-MDA: A Unified Framework for Long-Short Term Temporal Convolution and Mixed Data Augmentation in Skeleton-Based Action Recognition},
  author={Ding, Feng and Fu, Haisheng and Oraki, Soroush and Liang, Jie},
  journal={arXiv preprint arXiv:2509.14619},
  year={2025}
}

@inproceedings{stgcn,
  title={Spatial temporal graph convolutional networks for skeleton-based action recognition},
  author={Yan, Sijie and Xiong, Yuanjun and Lin, Dahua},
  booktitle={Proceedings of the AAAI conference on artificial intelligence},
  volume={32},
  year={2018}
}

@inproceedings{chen2021channelwise,
    title={Channel-wise Topology Refinement Graph Convolution for Skeleton-Based Action Recognition},
    author={Chen, Yuxin and Zhang, Ziqi and Yuan, Chunfeng and Li, Bing and Deng, Ying and Hu, Weiming},
    booktitle={Proceedings of the IEEE/CVF International Conference on Computer Vision},
    pages={13359--13368},
    year={2021}
  }

@InProceedings{Chi_2022_CVPR,
    author    = {Chi, Hyung-gun and Ha, Myoung Hoon and Chi, Seunggeun and Lee, Sang Wan and Huang, Qixing and Ramani, Karthik},
    title     = {InfoGCN: Representation Learning for Human Skeleton-Based Action Recognition},
    booktitle = {Proceedings of the IEEE/CVF Conference on Computer Vision and Pattern Recognition (CVPR)},
    month     = {June},
    year      = {2022},
    pages     = {20186-20196}
}

@misc{duan2022revisiting,
      title={Revisiting Skeleton-based Action Recognition}, 
      author={Haodong Duan and Yue Zhao and Kai Chen and Dahua Lin and Bo Dai},
      year={2022},
      eprint={2104.13586},
      archivePrefix={arXiv},
      primaryClass={cs.CV}
}

@article{ntu120,
   title={NTU RGB+D 120: A Large-Scale Benchmark for 3D Human Activity Understanding},
   volume={42},
   ISSN={1939-3539},
   number={10},
   journal={IEEE Transactions on Pattern Analysis and Machine Intelligence},
   publisher={Institute of Electrical and Electronics Engineers (IEEE)},
   author={Liu, Jun and Shahroudy, Amir and Perez, Mauricio and Wang, Gang and Duan, Ling-Yu and Kot, Alex C.},
   year={2020},
   month=oct, pages={2684–2701} }

@inproceedings{ntu,
  title={Ntu rgb+ d: A large scale dataset for 3d human activity analysis},
  author={Shahroudy, Amir and Liu, Jun and Ng, Tian-Tsong and Wang, Gang},
  booktitle={Proceedings of the IEEE conference on computer vision and pattern recognition},
  pages={1010--1019},
  year={2016}
}

@article{liu2025beyond,
  title={Beyond-Skeleton: Zero-shot Skeleton Action Recognition enhanced by supplementary RGB visual information},
  author={Liu, Hongjie and Niu, Yingchun and Zeng, Kun and Liu, Chun and Hu, Mengjie and Song, Qing},
  journal={Expert Systems with Applications},
  volume={273},
  pages={126814},
  year={2025},
  publisher={Elsevier}
}

@inproceedings{zhou2023zero,
  title={Zero-shot skeleton-based action recognition via mutual information estimation and maximization},
  author={Zhou, Yujie and Qiang, Wenwen and Rao, Anyi and Lin, Ning and Su, Bing and Wang, Jiaqi},
  booktitle={Proceedings of the 31st ACM international conference on multimedia},
  pages={5302--5310},
  year={2023}
}

@inproceedings{gupta2021syntactically,
  title={Syntactically guided generative embeddings for zero-shot skeleton action recognition},
  author={Gupta, Pranay and Sharma, Divyanshu and Sarvadevabhatla, Ravi Kiran},
  booktitle={2021 IEEE International Conference on Image Processing (ICIP)},
  pages={439--443},
  year={2021},
  organization={IEEE}
}

@inproceedings{li2023multi,
  title={Multi-semantic fusion model for generalized zero-shot skeleton-based action recognition},
  author={Li, Ming-Zhe and Jia, Zhen and Zhang, Zhang and Ma, Zhanyu and Wang, Liang},
  booktitle={International Conference on Image and Graphics},
  pages={68--80},
  year={2023},
  organization={Springer}
}

@inproceedings{cada,
  title={Generalized zero-and few-shot learning via aligned variational autoencoders},
  author={Schonfeld, Edgar and Ebrahimi, Sayna and Sinha, Samarth and Darrell, Trevor and Akata, Zeynep},
  booktitle={Proceedings of the IEEE/CVF conference on computer vision and pattern recognition},
  pages={8247--8255},
  year={2019}
}

@inproceedings{DiT,
  title={Scalable diffusion models with transformers},
  author={Peebles, William and Xie, Saining},
  booktitle={Proceedings of the IEEE/CVF international conference on computer vision},
  pages={4195--4205},
  year={2023}
}

@inproceedings{hubert2017learning,
  title={Learning robust visual-semantic embeddings},
  author={Hubert Tsai, Yao-Hung and Huang, Liang-Kang and Salakhutdinov, Ruslan},
  booktitle={Proceedings of the IEEE International conference on Computer Vision},
  pages={3571--3580},
  year={2017}
}

@inproceedings{wray2019fine,
  title={Fine-grained action retrieval through multiple parts-of-speech embeddings},
  author={Wray, Michael and Larlus, Diane and Csurka, Gabriela and Damen, Dima},
  booktitle={Proceedings of the IEEE/CVF international conference on computer vision},
  pages={450--459},
  year={2019}
}

@article{pytorch,
  title={Pytorch: An imperative style, high-performance deep learning library},
  author={Paszke, Adam and Gross, Sam and Massa, Francisco and Lerer, Adam and Bradbury, James and Chanan, Gregory and Killeen, Trevor and Lin, Zeming and Gimelshein, Natalia and Antiga, Luca and others},
  journal={Advances in neural information processing systems},
  volume={32},
  year={2019}
}

@misc{SkeletonFall,
  title = {Method and system for privacy-preserving fall detection},
  author={Ng, H. W. and \emph{et al.}},
  year={2021},
  note={{US} Patent 11179064}
}

@article{lecun2006tutorial,
  title={A tutorial on energy-based learning},
  author={LeCun, Yann and Chopra, Sumit and Hadsell, Raia and Ranzato, M and Huang, Fujie and others},
  journal={Predicting structured data},
  volume={1},
  number={0},
  year={2006}
}

@article{du2019implicit,
  title={Implicit generation and modeling with energy based models},
  author={Du, Yilun and Mordatch, Igor},
  journal={Advances in neural information processing systems},
  volume={32},
  year={2019}
}

@inproceedings{zhai2016deep,
  title={Deep structured energy based models for anomaly detection},
  author={Zhai, Shuangfei and Cheng, Yu and Lu, Weining and Zhang, Zhongfei},
  booktitle={International conference on machine learning},
  pages={1100--1109},
  year={2016},
  organization={PMLR}
}

@article{al2022energy,
  title={Energy-based learning for open-set classification in remote sensing imagery},
  author={Al Rahhal, Mohamad M and Bazi, Yakoub and Al-Dayil, Reham and Alwadei, Bashair M and Ammour, Nassim and Alajlan, Naif},
  journal={International Journal of Remote Sensing},
  volume={43},
  number={15-16},
  pages={6027--6037},
  year={2022},
  publisher={Taylor \& Francis}
}

@inproceedings{li2022energy,
  title={Energy-based models for continual learning},
  author={Li, Shuang and Du, Yilun and Van de Ven, Gido and Mordatch, Igor},
  booktitle={Conference on lifelong learning agents},
  pages={1--22},
  year={2022},
  organization={PMLR}
}

@article{kendall2017uncertainties,
  title={What uncertainties do we need in bayesian deep learning for computer vision?},
  author={Kendall, Alex and Gal, Yarin},
  journal={Advances in neural information processing systems},
  volume={30},
  year={2017}
}

@article{charpentier2020posterior,
  title={Posterior network: Uncertainty estimation without ood samples via density-based pseudo-counts},
  author={Charpentier, Bertrand and Z{\"u}gner, Daniel and G{\"u}nnemann, Stephan},
  journal={Advances in neural information processing systems},
  volume={33},
  pages={1356--1367},
  year={2020}
}

@inproceedings{stutz2020confidence,
  title={Confidence-calibrated adversarial training: Generalizing to unseen attacks},
  author={Stutz, David and Hein, Matthias and Schiele, Bernt},
  booktitle={International conference on machine learning},
  pages={9155--9166},
  year={2020},
  organization={PMLR}
}

@article{khan2024human,
  title={Human action recognition using fusion of multiview and deep features: an application to video surveillance},
  author={Khan, Muhammad Attique and Javed, Kashif and Khan, Sajid Ali and Saba, Tanzila and Habib, Usman and Khan, Junaid Ali and Abbasi, Aaqif Afzaal},
  journal={Multimedia tools and applications},
  volume={83},
  number={5},
  pages={14885--14911},
  year={2024},
  publisher={Springer}
}

@inproceedings{kamthe2018suspicious,
  title={Suspicious activity recognition in video surveillance system},
  author={Kamthe, UM and Patil, CG},
  booktitle={2018 Fourth international conference on computing communication control and automation (ICCUBEA)},
  pages={1--6},
  year={2018},
  organization={IEEE}
}

@article{kong2022human,
  title={Human action recognition and prediction: A survey},
  author={Kong, Yu and Fu, Yun},
  journal={International Journal of Computer Vision},
  volume={130},
  number={5},
  pages={1366--1401},
  year={2022},
  publisher={Springer}
}

@article{kumar2024survey,
  title={A survey on intelligent human action recognition techniques},
  author={Kumar, Rahul and Kumar, Shailender},
  journal={Multimedia Tools and Applications},
  volume={83},
  number={17},
  pages={52653--52709},
  year={2024},
  publisher={Springer}
}

@article{wang2023comprehensive,
  title={A comprehensive survey of rgb-based and skeleton-based human action recognition},
  author={Wang, Cailing and Yan, Jingjing},
  journal={IEEE Access},
  volume={11},
  pages={53880--53898},
  year={2023},
  publisher={IEEE}
}

@article{gammulle2023continuous,
  title={Continuous human action recognition for human-machine interaction: a review},
  author={Gammulle, Harshala and Ahmedt-Aristizabal, David and Denman, Simon and Tychsen-Smith, Lachlan and Petersson, Lars and Fookes, Clinton},
  journal={ACM Computing Surveys},
  volume={55},
  number={13s},
  pages={1--38},
  year={2023},
  publisher={ACM New York, NY}
}

@article{singh2019multi,
  title={Multi-view recognition system for human activity based on multiple features for video surveillance system},
  author={Singh, Roshan and Kushwaha, Alok Kumar Singh and Srivastava, Rajeev},
  journal={Multimedia Tools and Applications},
  volume={78},
  number={12},
  pages={17165--17196},
  year={2019},
  publisher={Springer}
}

@article{haria2017hand,
  title={Hand gesture recognition for human computer interaction},
  author={Haria, Aashni and Subramanian, Archanasri and Asokkumar, Nivedhitha and Poddar, Shristi and Nayak, Jyothi S},
  journal={Procedia computer science},
  volume={115},
  pages={367--374},
  year={2017},
  publisher={Elsevier}
}

@article{dibenedetto2025comparing,
  title={Comparing human pose estimation through deep learning approaches: An overview},
  author={Dibenedetto, Gaetano and Sotiropoulos, Stefanos and Polignano, Marco and Cavallo, Giuseppe and Lops, Pasquale},
  journal={Computer Vision and Image Understanding},
  pages={104297},
  year={2025},
  publisher={Elsevier}
}

@inproceedings{do2024skateformer,
  title={Skateformer: skeletal-temporal transformer for human action recognition},
  author={Do, Jeonghyeok and Kim, Munchurl},
  booktitle={European Conference on Computer Vision},
  pages={401--420},
  year={2024},
  organization={Springer}
}

@inproceedings{zhou2024blockgcn,
  title={Blockgcn: Redefine topology awareness for skeleton-based action recognition},
  author={Zhou, Yuxuan and Yan, Xudong and Cheng, Zhi-Qi and Yan, Yan and Dai, Qi and Hua, Xian-Sheng},
  booktitle={Proceedings of the IEEE/CVF Conference on Computer Vision and Pattern Recognition},
  pages={2049--2058},
  year={2024}
}

@article{ramirez2021fall,
  title={Fall detection and activity recognition using human skeleton features},
  author={Ramirez, Heilym and Velastin, Sergio A and Meza, Ignacio and Fabregas, Ernesto and Makris, Dimitrios and Farias, Gonzalo},
  journal={Ieee Access},
  volume={9},
  pages={33532--33542},
  year={2021},
  publisher={IEEE}
}
\end{document}